\documentclass[11pt]{article}

\usepackage[preprint]{acl}

\usepackage{times}
\usepackage{latexsym}

\usepackage[T1]{fontenc}
\usepackage[utf8]{inputenc}

\usepackage{microtype}

\usepackage{inconsolata}

\usepackage{graphicx}

\usepackage{booktabs}
\usepackage{amsmath}
\usepackage{amssymb}
\usepackage{multirow}
\usepackage{placeins}

\title{HydraQE: OSU's Submission for the IWSLT 2026 Speech Translation Metrics Shared Task}

\author{Kevin Krahn \and Eric Fosler-Lussier \\
        Dept. of Computer Science and Engineering \\
        The Ohio State University \\
        \texttt{\{krahn.6, fosler-lussier.1\}}@osu.edu }

\begin{document}
\maketitle
\begin{abstract}
We present HydraQE, our contribution to the IWSLT 2026 Speech Translation Metrics shared task.
HydraQE is an end-to-end, reference-free quality estimation (QE) system for speech translation built on a Qwen3-ASR backbone, which accepts source audio and a translation hypothesis as joint input. Hidden states from all backbone layers are combined via a learnable sparsemax scalar mix, then re-encoded by a lightweight bidirectional Transformer to enable full cross-modal interaction prior to pooling into a shared embedding. Three independent prediction heads are trained on complementary supervision signals: human direct assessment (DA) annotations, MetricX-24 pseudo-labels, and xCOMET pseudo-labels. To address the scarcity of human-annotated data, we train on a combination of synthetically corrupted examples and silver pseudo-labeled machine translation outputs, using a curriculum that begins on synthetic and silver data and gradually shifts toward human-annotated examples. HydraQE outperforms cascaded text-based baselines and prior direct speech QE systems, demonstrating that end-to-end speech translation QE is competitive with cascaded approaches.

\end{abstract}

\section{Introduction}
Quality estimation (QE) systems predict the quality of a machine translation given only the source and a candidate translation. While the ratings produced by recent QE systems correlate well with human judgments \citep{rei-etal-2023-scaling, juraska-etal-2024-metricx}, their emphasis has been on text translation, requiring a combination of automatic speech recognition (ASR) and text-QE for evaluating speech translations. Such a cascaded approach is inefficient and can pass ASR errors into the text-QE system, compounding errors and limiting performance.

\begin{figure}[t!]
    \centering
    \includegraphics[width=\linewidth]{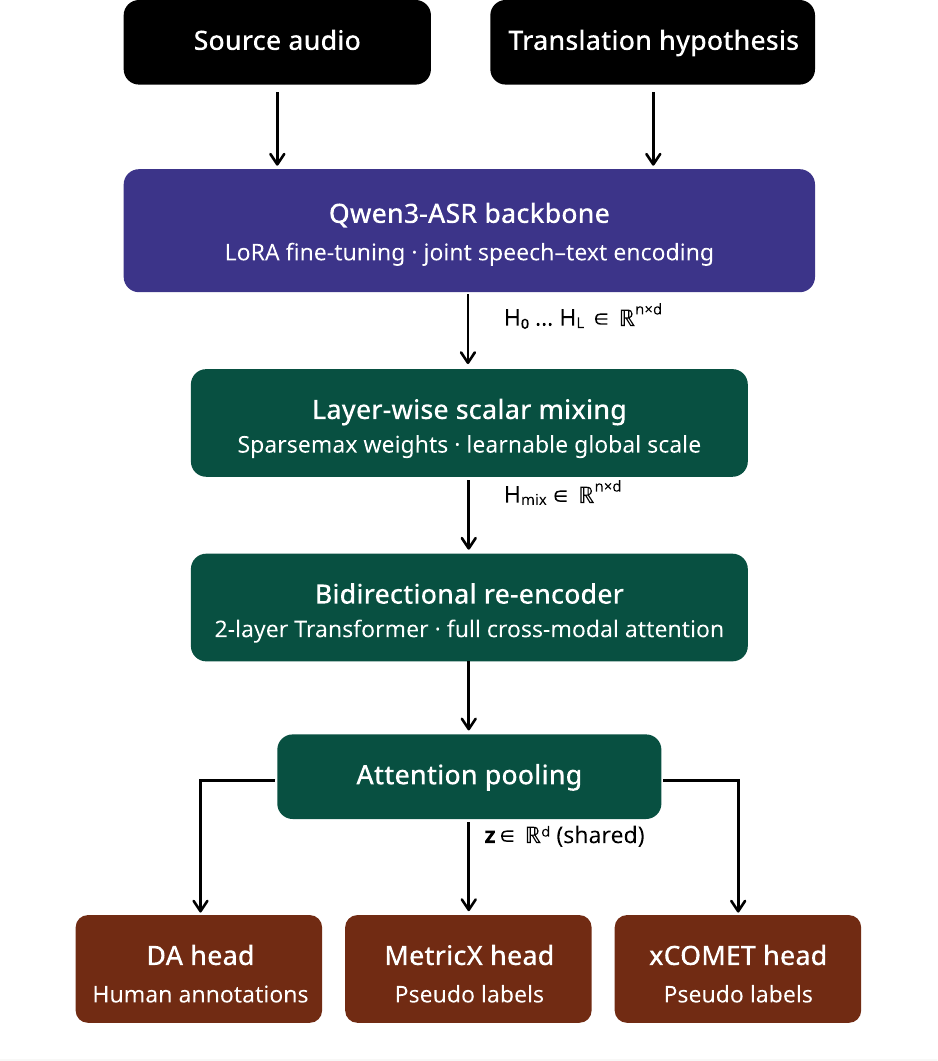}
    \caption{HydraQE architecture: Source audio and translation hypothesis are fed into the backbone model, followed by a scalar mixing module and bidirectional re-encoder. The token-wise embeddings are pooled and fed into multiple prediction heads.}
    \label{fig:hydraqe-diagram}
\end{figure}

In this paper, we present our submission to the IWSLT 2026 Speech Translation Metrics shared task \citep{adelani-etal-2026-iwslt}, focusing on an end-to-end speech QE system for reference-free evaluation of speech translation quality. Our design incorporates the Qwen3-ASR \citep{shi2026qwen3asrtechnicalreport} model as our pretrained speech and text backbone along with multiple prediction heads trained on complementary supervision signals: (1) human direct assessment (DA) annotations from the official training data, (2) silver pseudo-labels generated by text metric models applied to machine-translated speech translation outputs, and (3) synthetically corrupted examples derived from speech translation datasets. Because human-annotated data is scarce, we employ a curriculum sampling strategy that begins training on synthetic and pseudo-labeled data and shifts toward human-labeled data.

Our system outperforms cascaded text-based baselines and prior speech-based systems across language pairs, on both segment-level and system-level evaluations. Our results indicate that end-to-end speech translation QE systems can outperform cascaded approaches while being simpler and more efficient.

\section{Related Work}
Reference-free quality estimation has advanced rapidly with
the adoption of pre-trained cross-lingual encoders as backbones. COMET
\citep{rei-etal-2020-comet} established the paradigm of fine-tuning a
multilingual encoder on human quality annotations, and subsequent work scaled
this to larger models, larger datasets, and word-level annotations
\citep{rei-etal-2022-cometkiwi, rei-etal-2023-scaling, guerreiro-etal-2024-xcomet}.
MetricX-24 \citep{juraska-etal-2024-metricx} demonstrated that large-scale
synthetic data constructed by systematically corrupting reference translations
provides a strong training signal for rare failure modes, and further
introduced a unified hybrid architecture capable of reference-based and
reference-free evaluation within a single model trained on a mixture of DA
and MQM ratings. LLM-based approaches such as GEMBA-MQM
\citep{kocmi-federmann-2023-gemba} have demonstrated strong zero-shot QE
capabilities using generative models, especially at the system-level, but
still fall short of trained metrics at the segment level \citep{lavie-etal-2025-findings}.

Evaluation of speech translation has historically followed MT evaluation
practice by applying text metrics such as BLEU \citep{papineni-etal-2002-bleu}, COMET \citep{rei-etal-2020-comet}, and BLEURT \citep{sellam-etal-2020-bleurt} to system output \citep{agostinelli-etal-2025-findings}, but several recent systems support direct speech QE. The BLASER family of models operates in a shared cross-lingual and cross-modal speech--text embedding space, allowing for superior transfer across modalities and languages where labeled data is unavailable. The original BLASER \citep{chen-etal-2023-blaser} introduced a speech translation evaluation metric based on LASER \citep{jones-etal-2021-massively} sentence embeddings, and BLASER 2.0 \citep{dale-costa-jussa-2024-blaser} improved upon this
with multi-modal SONAR embeddings \citep{duquenne2023sonarsentencelevelmultimodallanguageagnostic}, using a multilingual speech encoder trained in a teacher--student fashion to make the speech encoder match the SONAR text encoder sentence embeddings.
BLASER 3.0 \citep{omnilingualmtteam2026omnilingualmtmachinetranslation} adopts a
multi-head architecture with separate prediction heads per training signal,
which inspires our design, although it is not publicly available for testing.
SpeechQE \citep{han-etal-2024-speechqe} is
a model designed specifically for speech translation QE, combining a pretrained
speech encoder and text LLM trained on a QE task.

The effectiveness of direct speech QE compared to ASR-based approaches has been mixed; text metrics such as
xCOMET and MetricX combined with ASR frequently achieve stronger correlation with human judgments
than BLASER 2.0 \citep{han-etal-2024-speechqe, cettolo2026evaluatespeechtranslationsourceaware},
while SpeechQE matches or slightly outperforms cascaded ASR approaches in some
scenarios \citep{han-etal-2024-speechqe}.

\begin{table}[t]
   \centering
   \small
   \begin{tabular}{llr}
   \toprule
   \textbf{Source} & \textbf{Language pairs} & \textbf{Segments} \\
   \midrule
   IWSLT 2023     & en$\rightarrow$de, en$\rightarrow$ja, en$\rightarrow$zh                 & 12,480 \\
   WMT 2024       & en$\rightarrow$cs, en$\rightarrow$es, en$\rightarrow$is, & \\
                  & en$\rightarrow$ja, en$\rightarrow$ru, en$\rightarrow$uk, & \\
                  & en$\rightarrow$zh              & 10,193 \\
   WMT 2025       & cs$\rightarrow$de, cs$\rightarrow$uk, en$\rightarrow$cs, & \\
                  & en$\rightarrow$is, en$\rightarrow$ja, en$\rightarrow$ru, & \\
                  & en$\rightarrow$uk, en$\rightarrow$zh & 5,836 \\
   \midrule
   \textbf{Total} &                                              & \textbf{28,509} \\
   \bottomrule
   \end{tabular}
    \caption{Official training data segments after filtering to the 8 most-frequent target
     languages (\textit{zh, ja, de, uk, cs, ru, is, es}).}
   \label{tab:official-data}
\end{table}

\begin{table*}[!t]
   \centering
   \small
   \begin{tabular}{p{3cm} p{9cm} p{1.5cm}}
   \toprule
   \textbf{Corruption} & \textbf{Description} & \textbf{Score} \\
   \midrule
   \texttt{Undertranslation}
   & Randomly chosen sentence removed if segment is multi-sentence; otherwise, truncate $30$--$60\%$ of words.
   & $[0.15,0.5]$ \\

   \texttt{Corruption}
   & Replace a random $40$--$60\%$ of tokens with random vocabulary tokens from the same language.
   & $[0.2,0.3]$ \\

   \texttt{Unrelated}
   & Replace candidate translation with randomly chosen segment of similar length in the same language.
   & $0.0$\\

   \texttt{Gibberish}
   & Sample random vocabulary tokens to produce a sequence with the same length as the candidate.
   & $0.0$ \\

   \texttt{Identity (Gold)}
   & Use the reference translation.
   & $1.0$ \\
   \bottomrule
   \end{tabular}
   \caption{Synthetic corruption approaches and score targets.}
   \label{tab:synthetic-corruptions}
\end{table*}

\section{Methods}
\subsection{Dataset}
\paragraph{Official shared task data}
We use the official IWSLT 2026 shared task dataset,
which aggregates direct assessment (DA) annotations from IWSLT 2023 and WMT 2024--2025.
We filter the dataset to the 8 most-frequent target languages for training (see Table \ref{tab:official-data}).
We use the provided development set, consisting of human annotations of IWSLT 2025
ACL Talks, for all model selection and hyperparameter tuning. We convert all training labels
into the 0--1 range for consistency across sources.

\paragraph{Synthetic data}
To generate additional training samples, we apply
several corruptions to the reference translations (see Table \ref{tab:synthetic-corruptions})
in CoVoST~2\citep{DBLP:conf/interspeech/WangWGP21} and FLEURS
\citep{DBLP:conf/slt/ConneauMKZADRRB22}, inspired by the synthetic data
construction procedure of MetricX \citep{juraska-etal-2023-metricx, juraska-etal-2024-metricx}.
The \texttt{Unrelated}, \texttt{Gibberish}, and \texttt{Identity} samples have fixed scores, while the score targets for \texttt{Undertranslation} and \texttt{Corruption} are derived as the proportion of the character length left unmodified after corruption, then divided by two to account for the disfluency incurred by the corruptions in addition to semantic changes.
These synthetic examples primarily cover the low end of the score range for several reasons: (1) our QE datasets are biased toward high scores, so we attempt to partially balance the distribution; (2) it is difficult to procedurally generate useful scores for small corruptions, as changing one or two words could either drastically alter meaning or have little effect; (3) metric models struggle to detect undertranslations and fluent but unrelated translations \citep{juraska-etal-2024-metricx}, so we target these failures modes directly.

\paragraph{Silver data}
We also construct a large set of \emph{silver} examples by running MT
systems over CoVoST~2 and FLEURS English source segments to obtain
machine-generated translation hypotheses for en$\rightarrow$de and en$\rightarrow$zh
language pairs, then pseudo-labelling them
with MetricX-24-XXL \citep{juraska-etal-2024-metricx} and xCOMET-XXL \citep{guerreiro-etal-2024-xcomet} using the ground-truth reference translations as additional input.
We transform the outputs of the MetricX-24 model into the $[0,1]$ range: $(25-\text{score})/25$.
Training on these silver labels serves to distill these large text models
into our speech QE model.
We generate hypotheses using multiple MT systems: SeamlessM4T
\citep{communication2023seamlessm4tmassivelymultilingual}, which
translates directly from the source audio; and NLLB
\citep{nllbteam2022languageleftbehindscaling} and Hunyuan-MT-7B
\citep{zheng2025hunyuanmttechnicalreport}, which translate from the gold
transcripts. These systems vary widely in model size, training procedure,
and architecture, thus diversifying hypothesis quality and style.

\paragraph{TTS-augmented text QE data}
Human-labeled data for speech translation quality is limited. To further increase training coverage, we incorporate human-labeled text data from the WMT 2022 Quality Estimation shared task \citep{freitag-etal-2022-results}, which provides source--hypothesis pairs with human DA scores for the en→de and en→zh language pairs. We synthesize speech using Qwen3-TTS \citep{hu2026qwen3ttstechnicalreport}, allowing these examples to be used for speech-input QE training.

\subsection{Architecture}

The overall architecture for the system is shown in Figure \ref{fig:hydraqe-diagram}; a breakdown
of each component follows.

\paragraph{Backbone}
We use Qwen3-ASR-1.7B \citep{shi2026qwen3asrtechnicalreport} as our backbone encoder.
Qwen3-ASR supports speech and text inputs in 52 languages and dialects, allowing us to feed the
source audio and the translation hypothesis as a single forced input and produce
contextualized representations of both modalities jointly. We fine-tune the
backbone using Low-Rank Adaptation (LoRA) \citep{DBLP:conf/iclr/HuSWALWWC22}, which inserts
trainable low-rank matrices into the attention projections while keeping the
original weights frozen, substantially reducing the number of trainable
parameters and mitigating overfitting on the relatively small human-annotated
training set. We additionally freeze the speech encoder parameters, as
updating them degrades performance, with or without LoRA.
Although this model was fine-tuned for ASR and not translation tasks,
we find that using the translation hypothesis as forced input works well in
practice, likely due to the rich pretrained capabilities of the base Qwen3 model.

\paragraph{Layer-wise scalar mixing}
Rather than using only the final hidden layer, we extract hidden states from
all $L+1$ layers of the backbone.  For an input of $n$ tokens, layer $\ell$
produces $H_\ell \in \mathbb{R}^{n \times d}$.  Following
\citet{rei-etal-2023-scaling}, we combine these with a scalar mix module that
learns a task-specific weighted average across layers:
\begin{equation}
   H_{\mathrm{mix}} = \lambda \sum_{\ell=0}^{L} \beta_\ell\, H_\ell ,
\end{equation}
where $\lambda \in \mathbb{R}$ is a learnable global scale and
$\boldsymbol{\beta} \in \Delta^{L}$ is a layer-importance distribution
parameterised with \textsc{sparsemax} \citep{DBLP:conf/icml/MartinsA16}.
\textsc{Sparsemax} is preferred over \textsc{softmax} because it produces
sparse weights, effectively selecting a small subset of layers rather than
diffusely averaging all of them.

\paragraph{Bidirectional re-encoding}
Because Qwen3-ASR is a causal language model, each token in $H_{\mathrm{mix}}$
attends only to its left context.  Consequently, early speech tokens have no
direct access to the translation hypothesis and vice versa.  To enable full
cross-modal interaction, we pass $H_{\mathrm{mix}}$ through a lightweight
two-layer bidirectional Transformer encoder.  This re-encoding step is
inexpensive relative to the backbone but improves the quality of
the pooled representation by allowing every token to attend to the full
source--hypothesis context. A final attention pooling layer then reduces the
variable-length sequence into a single fixed-size embedding $\mathbf{z} \in
\mathbb{R}^{d}$, which is shared across all prediction heads.

\paragraph{Multiple scoring heads}
Inspired by BLASER 3.0 \citep{omnilingualmtteam2026omnilingualmtmachinetranslation},  we attach multiple independent prediction heads to $\mathbf{z}$, one per
training signal.  Unlike BLASER 3.0, which uses a single linear projection per
head, each of our heads is a two-layer feed-forward network with a \textsc{Tanh}
activation, producing a scalar quality score.  The
additional capacity allows each head to model the distributional characteristics
of its supervision source (human DA labels, MetricX pseudo-labels, and xCOMET
pseudo-labels) while limiting interference with the shared representation. The pseudo-label
heads effectively distill the reference-based metrics into a reference-free metric.

\subsection{Training Procedure}

\paragraph{Multi-head loss}
Each training example is routed to its corresponding prediction head based on data source: human DA annotations are passed to the DA head, and MetricX and xCOMET pseudo-labeled examples are passed to their respective silver heads. The total loss is the sum of the mean squared error (MSE) between the predicted and target scores, with the DA head weighted separately:
\begin{equation}
L_{total} = L_{\text{MetricX}} + L_{\text{xCOMET}} + \lambda_{DA} \cdot L_{\text{DA}}
\end{equation}
We find that $\lambda_{DA}=1.5$ works well in practice; we tried values in $\{1.0, 1.5, 2.0\}$ and observed only marginal differences, with $1.5$ performing slightly better on the development set.
Gradients flow through the active head and the shared backbone, so the backbone is updated by all data sources while each head specializes to its own distribution.

\begin{table*}[t]
\centering
\small
\begin{tabular}{l l ccc ccc ccc ccc}
\toprule
& & \multicolumn{6}{c}{\textbf{Dev}} & \multicolumn{6}{c}{\textbf{Test} (scored by organizers)} \\
\cmidrule(lr){3-8} \cmidrule(lr){9-14}
& & \multicolumn{3}{c}{\textbf{Segment-level} ($\tau$)}
  & \multicolumn{3}{c}{\textbf{System-level}}
  & \multicolumn{3}{c}{\textbf{Segment-level} ($\tau$)}
  & \multicolumn{3}{c}{\textbf{System-level}} \\
\cmidrule(lr){3-5} \cmidrule(lr){6-8}
\cmidrule(lr){9-11} \cmidrule(lr){12-14}
\textbf{System}
& & \textbf{de} & \textbf{zh} & \textbf{Avg}
  & \textbf{de} & \textbf{zh} & \textbf{Avg}
  & \textbf{de} & \textbf{zh} & \textbf{Avg}
  & \textbf{de} & \textbf{zh} & \textbf{Avg} \\
\midrule

COMET (partial)\textsuperscript{$\dagger$}
  & & 11.3 & 12.0 & 11.6 & 44.4 & 68.7 & 56.6
  & 13.1 & 14.8 & 14.0 & 85.7 & 78.2 & 81.9 \\

COMET\textsuperscript{$\dagger$}
  & & 32.6 & 36.5 & 34.6 & 86.2 & 92.6 & 89.4
  & - & - & - & - & - & - \\

SpeechQE\textsuperscript{$\dagger$}
  & & 26.6 & 31.8 & 29.2 & 78.6 & 93.4 & 86.0
  & 22.1 & 23.3 & 22.7 & 89.8 & 96.1 & 92.9 \\

BLASER 2.0 QE\textsuperscript{$\dagger$}
  & & 22.0 & 26.8 & 24.4 & 86.0 & 67.7 & 76.9
  & 18.5 & 20.3 & 19.4 & 88.8 & 88.2 & 88.5 \\

CometKiwi\textsuperscript{$\dagger$}\textsuperscript{*}
  & & 31.4 & 35.9 & 33.7 & 91.6 & 34.9 & 63.3
  & 23.7 & 33.3 & 28.5 & 93.1 & 97.2 & 95.2 \\

MetricX-24-XXL QE
  & & 29.7 & 31.2 & 30.5 & 99.6 & \textbf{91.3} & \textbf{95.5}
  & - & - & - & - & - & - \\

xCOMET-XXL QE
  & & 30.4 & 35.5 & 33.0 & 87.7 & 45.8 & 66.8
  & - & - & - & - & - & - \\

\midrule

HydraQE (DA head)
  & & 32.5 & \textbf{39.9} & \textbf{36.2} & \textbf{99.7} & 69.2 & 84.5
  & 25.3 & \textbf{33.6} & 29.5 & 94.7 & 98.1 & 96.4 \\

HydraQE (MetricX head)
  & & 31.8 & 39.3 & 35.6 & 98.8 & 90.1 & 94.5
  & 24.3 & 32.9 & 28.6 & 94.2 & 97.1 & 95.7 \\

HydraQE (xCOMET head)
  & & \textbf{32.7} & 37.1 & 34.9 & 96.9 & 34.6 & 65.7
  & \textbf{26.7} & 32.9 & \textbf{29.8} & 94.8 & \textbf{98.7} & 96.7 \\

HydraQE (all heads avg.)
  & & 32.0 & 38.7 & 35.4 & 99.5 & 56.0 & 77.7
  & 25.3 & 33.3 & 29.3 & \textbf{95.1} & 98.4 & \textbf{96.8} \\

HydraQE (primary)
  & & 32.1 & 39.8 & 36.0 & 99.4 & 79.3 & 89.4
  & 24.8 & 33.5 & 29.1 & 94.5 & 97.9 & 96.2 \\

\bottomrule
\end{tabular}
\caption{Segment-level Kendall-Tau ($\tau$)($\uparrow$) and System-level Soft Pairwise Accuracy ($\uparrow$) of HydraQE and baselines on the IWSLT 2026 Metrics Shared Task development and test sets, with $en$ as the source language and \textit{de} and \textit{zh} as the target languages.
The primary submission is a weighted average of the DA and MetricX heads.
$\dagger$~denotes official shared task baselines; text-based baselines use gold transcripts provided with the dataset.
*~For CometKiwi, we report dev results using the XXL variant; on the test set we report the organizers' results.
}
\label{tab:main-results}
\end{table*}

\paragraph{Curriculum sampling}
Because human DA annotations are scarce relative to the synthetic and
silver data, we train with a curriculum that gradually shifts the
sampling distribution toward official data.
We control the DA/non-DA split with a mixing coefficient $\alpha$,
which remains at $0$ for the first 10,000 steps (a warm-up phase on
synthetic and silver data), then increases linearly to $0.09$ over the
next 20,000 steps. At each step, we sample from the official DA pool
with probability $\alpha$ and from the synthetic--silver pool with
probability $1 - \alpha$.

Within the non-DA pool, a second coefficient $\beta$ controls the
synthetic/silver balance: synthetic examples are drawn with probability
$\beta$ and silver examples with probability $1 - \beta$. $\beta$
decreases linearly from $1.0$ to $0.005$ over the first 5,000 steps.
The synthetic data thus contributes
heavily at first but quickly tapers off into a small
but consistent signal. All schedule values
($\alpha$, $\beta$, and the step counts) were tuned
on the development set.

To avoid the model overfitting to common sequence lengths and
score ranges, silver and synthetic examples are drawn via stratified
sampling: we maintain five equal-mass bins along the sequence-length
axis and, for silver data, five equal-mass bins along the pseudo-label
score axis, sampling uniformly across bins before sampling within them.
Additionally, we balance languages within each
pool so each language pair contributes equally.

Synthetic examples are routed to the silver heads but are
deliberately withheld from the DA head. Routing synthetic examples to
the DA head slightly degrades performance on human DA annotations,
likely because the deterministic corruption-based scores used for
synthetic data do not match the distributional characteristics of human
judgments.
%Routing noisy synthetic signal into this head would distort the learned score distribution.

\paragraph{Checkpoint selection}
We train the model for 70,000 steps with a batch size of 8, using the AdamW optimizer.
We evaluate performance on the dev set every 1,000 steps.
Through experimentation with the augmented WMT 2022 QE en$\rightarrow$de and en$\rightarrow$zh TTS data,
we find that even when sampled with low probability the inclusion of this data slightly
hurt performance, and therefore exclude it from our final training run.

Based on dev scores, a tradeoff exists in the choice of
prediction head: the DA head achieves the highest
segment-level scores, while the MetricX head achieves the highest system-level scores.
To balance between these tasks for our primary model, we independently choose the checkpoint
with the best segment-level scores
per head and combine their predictions using the following sum:
\begin{equation}
\text{Score} = \frac{3}{4}\text{Head}_{\text{DA}} + \frac{1}{4}\text{Head}_{\text{MetricX}}.
\end{equation}
The higher weight for the DA head is chosen to prioritize segment-level over system-level scores.

\begin{figure*}[t]
    \centering
    \includegraphics[width=\linewidth]{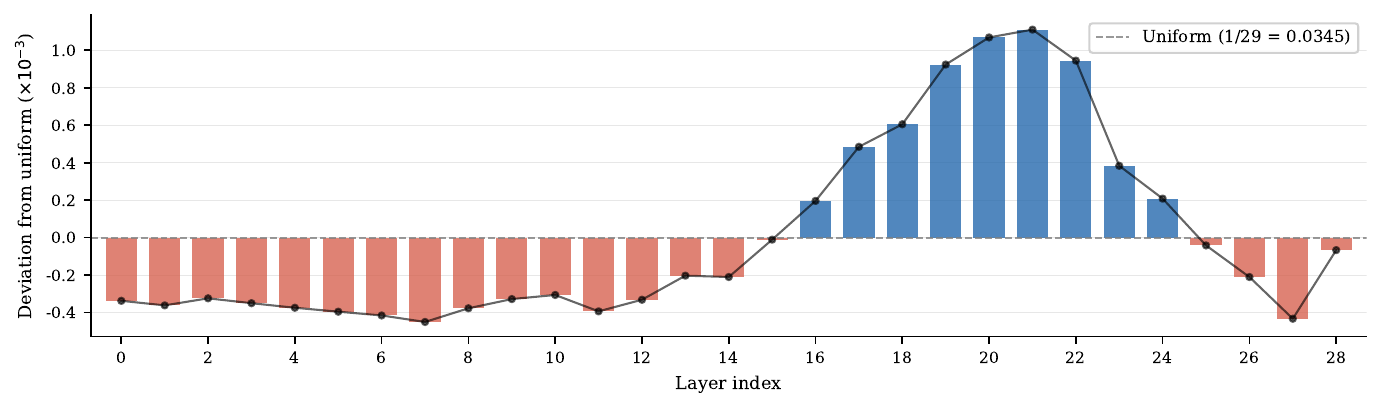}
    \caption{Learned scalar-mix weights of layer-wise mixing module, shown as deviation from uniform weighting ($1/29 \approx 0.034$). The model up-weights upper-middle layers (16--24), consistent with those layers carrying richer semantic representations, while suppressing the earliest and topmost layers.}
    \label{fig:scalar_mix}
\end{figure*}

\begin{figure*}[t]
    \centering
    \includegraphics[width=\linewidth]{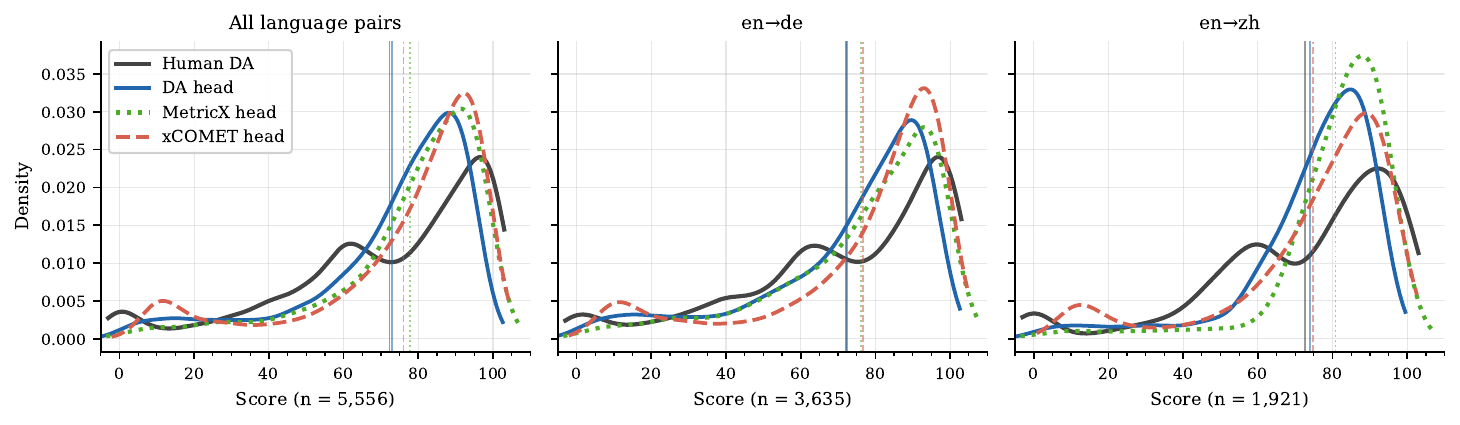}
    \caption{Distributions of human scores and predicted scores from each HydraQE head on the IWSLT 2026 dev set.}
    \label{fig:head_distributions}
\end{figure*}

\section{Evaluation}

\subsection{Main Results}
Table~\ref{tab:main-results} reports segment-level Kendall-$\tau$ and
system-level Soft Pairwise Accuracy for all HydraQE configurations and baselines.
We report our results on the development set and the IWSLT organizers' official results
on the test set.

On segment-level evaluation, HydraQE achieves the strongest
performance across both the dev and test sets, reaching $\tau = 33.6$ on en$\rightarrow$zh on the test set using the DA head, and
$29.8$ averaged across language pairs using the xCOMET head, outperforming all baselines
including the strongest text-based baseline, CometKiwi ($28.5$), which has
access to gold transcripts.  The fact that an end-to-end speech QE
system exceeds a text-based cascade that uses gold transcripts suggests
that the joint audio--hypothesis encoding
captures quality-relevant information beyond what is available in
transcripts alone. The speech-native baselines, SpeechQE and BLASER
2.0 QE, trail substantially at $22.7$ and $19.4$ respectively,
confirming that prior direct speech QE systems leave considerable room
for improvement. The poor performance of BLASER 2.0 is possibly due to
it being trained on XSTS annotations rather than DA annotations.

System-level results present a different picture. On the dev set, MetricX-24-XXL
achieves the highest system-level average ($95.5$), driven by a near-perfect
score on en$\rightarrow$de ($99.6$).  The MetricX head of HydraQE closely
tracks this behavior ($94.5$), while the DA head, despite its strong
segment-level performance, drops to $84.5$ due to weaker performance on
en$\rightarrow$zh ($69.2$). The xCOMET head performs poorly on
en$\rightarrow$zh ($34.6$), lower than the teacher metric xCOMET ($45.8$).
%The poor system-level performance of the newer xCOMET and CometKiwi models is surprising, as they would be expected to outperform the older COMET model.
These dev set results motivated
us to blend the DA and MetricX heads for our primary submission, trading a small amount of
segment-level performance relative to the DA head alone in exchange for a substantially
improved system-level average. However, on the test
set all heads achieve excellent average system scores ($>95$).
In fact, the worst HydraQE head on the dev set, xCOMET, is the best individual head on the test set.
We also note that the system-level dev scores
oscillated dramatically during training, making these scores somewhat untrustworthy.
The ensemble of all heads yielded the highest average system test score ($96.8$),
surpassing the xCOMET head ($96.7$) by a narrow margin.

The primary submission combines the DA and MetricX heads, based on dev set evidence.
The test results show this to be a reasonable, but not optimal choice, not beating an
individual head in any category but also outperforming the worst head in all categories. For
comparison with other submissions, we refer the reader to the official evaluation results
for the IWSLT metrics shared task \citep{adelani-etal-2026-iwslt}.

\subsection{Analysis}

\paragraph{Layer-wise scalar mixing}
Figure~\ref{fig:scalar_mix} shows that the learned scalar mix concentrates
weight in the upper-middle layers (16--24) of the backbone, with near-zero
weight assigned to the lowest and highest layers.  This is consistent with
probing studies of transformer language models, which find that the topmost layers
tend to specialize on token prediction rather than general-purpose meaning
\citep{jawahar-etal-2019-bert, tenney-etal-2019-bert, liu-etal-2024-fantastic}.
The pattern suggests
that the scalar mix is exploiting the backbone as a semantic encoder rather
than relying on its ASR-tuned output layer, which is encouraging given that
the model was not originally trained for translation quality assessment.

\paragraph{Score distributions by head}
Figure~\ref{fig:head_distributions} reveals qualitatively different output
distributions across the three prediction heads. The human DA labels are
concentrated overall toward the high-end, but show distinct, smaller concentrations in the middle
and low end of the score range, reflecting annotators' tendency to cluster
translations scores rather than using the entire score range. The DA head
trained on human labels produces a smoother, unimodal distribution
concentrated toward the high end. The two pseudo-label heads similarly concentrate
toward the high end, consistent with prior work showing automatic metrics to be
overly optimistic and generous with scores for low quality translations
\citep{zouhar-etal-2024-pitfalls}. The inclusion of synthetic data covering
the extreme low end was unsuccessful at mitigating this issue.

The two pseudo-label heads also exhibit language-specific behavior: the
MetricX head concentrates more density at the high end for en$\rightarrow$de,
while the xCOMET head does so for en$\rightarrow$zh.  This asymmetry likely
reflects differing calibration of the two teacher metrics across languages
and may partly explain the complementarity between the DA and MetricX
heads observed in the dev results.

\section{Conclusion}
In this work we presented HydraQE, an end-to-end reference-free quality estimation
system for speech translation built on a Qwen3-ASR backbone.  By jointly
encoding source audio and translation hypothesis and training multiple
prediction heads on complementary supervision signals via curriculum
sampling, HydraQE achieves stronger segment-level correlation with human
judgments on the IWSLT 2026 Metrics Shared Task test set
compared to both cascaded text-based baselines
and prior direct speech QE systems.

A key finding is that the choice of prediction head involves a meaningful
tradeoff in per-language performance as well as segment vs system-level performance.
No head dominates all metrics simultaneously.  Our primary submission attempts to balance
segment and system-level performance by combining two heads based on dev performance, but our
post-hoc test set analysis showed no benefit over the single best head or an average over all heads.

Several directions remain open for future work.  The curriculum sampling
schedule and head weighting were tuned on a single development set
covering only two language pairs; evaluation across a broader set
of source and target languages would strengthen confidence in these design
choices.  The exclusion of TTS-augmented text QE data, which marginally
hurt development performance, warrants further investigation, as
synthesized speech may introduce domain mismatch that targeted data
augmentation could address.

\section*{Acknowledgments}
This work was carried out in partnership with the ETEN Innovation Lab. We are grateful to Joel Mathew for his guidance and support throughout this project, and to the members of the OSU SLaTe Lab for many helpful discussions.

\bibliography{anthology-1,anthology-2,custom}

\vspace{1cm}
\appendix

\section{Hyperparameters}
\label{sec:hyperparameters}
\begin{table}[h]
\centering
\small
\begin{tabular}{ll}
\toprule
\textbf{Hyperparameter} & \textbf{Value} \\
\midrule
\multicolumn{2}{l}{\textit{Optimization}} \\
Optimizer                        & AdamW \\
AdamW $\beta_1$, $\beta_2$       & 0.9, 0.999 \\
Learning rate (backbone)         & 5e-6 \\
Learning rate (new params)       & 2e-5 \\
Weight decay                     & 0.05 \\
Warmup steps                     & 5{,}000\\
Batch size                       & 8 \\
Training steps                   & 70{,}000 \\
Evaluation interval              & 1{,}000 steps \\
Dropout                          & 0.1 \\
\midrule
LoRA Rank                        & 32 \\
LoRA Alpha                       & 64 \\
LoRA Dropout                     & 0.1 \\
\midrule
\multicolumn{2}{l}{\textit{Re-encoder}} \\
Layers                           & 2 \\
Hidden size                      & 1024 \\
Attention heads                  & 8 \\
Activation                       & \textsc{SwiGLU}\\
\midrule
\multicolumn{2}{l}{\textit{Prediction Heads}} \\
Hidden sizes                     & 3072, 1024 \\
Activation                       & \textsc{Tanh} \\
\midrule
\multicolumn{2}{l}{\textit{Curriculum sampling}} \\
DA warm-up duration              & 10{,}000 steps \\
DA ramp duration                 & 20{,}000 steps \\
DA mixing coefficient $\alpha$ (final) & 0.09 \\
Synthetic ramp duration          & 5{,}000 steps \\
Synthetic coefficient $\beta$ (start) & 1.0 \\
Synthetic coefficient $\beta$ (final) & 0.005 \\
\bottomrule
\end{tabular}
\caption{Hyperparameters used for the final HydraQE model.}
\label{tab:hyperparameters}
\end{table}

\end{document}